# A SURVEY OF THE TRENDS IN FACIAL AND EXPRESSION RECOGNITION DATABASES AND METHODS

Sohini Roychowdhury[1] and Michelle Emmons[2]

[1]Department of Electrical Engineering, University of Washington, Bothell,
roych@uw.edu
[2]Department of Electrical Engineering, University of Washington, Bothell
memmons1442@gmail.com

*ABSTRACT*

*Automated facial identification and facial expression recognition have been topics of active research over the past few decades. Facial and expression recognition find applications in human-computer interfaces, subject tracking, real-time security surveillance systems and social networking. Several holistic and geometric methods have been developed to identify faces and expressions using public and local facial image databases. In this work we present the evolution in facial image data sets and the methodologies for facial identification and recognition of expressions such as anger, sadness, happiness, disgust, fear and surprise. We observe that most of the earlier methods for facial and expression recognition aimed at improving the recognition rates for facial feature-based methods using static images. However, the recent methodologies have shifted focus towards robust implementation of facial/expression recognition from large image databases that vary with space (gathered from the internet) and time (video recordings). The evolution trends in databases and methodologies for facial and expression recognition can be useful for assessing the next-generation topics that may have applications in security systems or personal identification systems that involve "Quantitative face" assessments.*

## 1. INTRODUCTION

Facial recognition has been a problem of interest since the early 1960's [1] when semi-automated methods were developed to manually locate facial features (eyes, ears, nose, mouth) followed by calculation of distances to reference points for recognition tasks. Over the next few decades facial feature detection and calculation of distances between the features and reference points became a favoured practice [2]. The first work to stray from the concept of facial feature extraction and to look at the holistic information in the residual of Eigen-faces was introduced in 1991 [3]. Since then several methodologies have been developed that analyse certain facial features or the Eigen-decomposition of faces or combine both strategies [4]. The increasing number of methods for facial and expression recognition methods spurred the development of several databases that can be used for benchmarking and comparative assessment of the methodologies. Over the years, specific trends have emerged from the facial databases and the methodologies that demonstrate certain technological and computational dependencies. In this paper, we analyse past trends in the facial recognition databases and methods to assess some of the future trends in the domain of facial recognition and expression analysis.

The primary challenges associated with facial and expression recognition include variations in the following factors: lighting, pose, imaging modalities, occlusions and expressions. Other limiting factors include gender, age and complexion. Additionally, variations in image qualities due to data compression formats, image blurriness and variations in imaging angles impose additional constraints for automated facial and expression recognition algorithms. Two-dimensional (2D) facial recognition algorithms such as the ones in [4-8] have shown high success rate in a controlled environment, but in an uncontrolled setting their performance has been shown to drastically decrease [9]. Till date, 2D facial recognition techniques have been explored for longer than three-dimensional (3D) algorithms; nonetheless the 3D facial recognition methods have been found to be more effective in controlled and uncontrolled settings [10].

In the present day facial detection and expression recognition finds many real-time applications such as: design of human-computer interfaces, to real-time video surveillance systems, security systems [11] and expression tagging on social media [12]. With the changes in computational technologies, two categories of automated facial recognition algorithms emerged. While the first category of holistic algorithms analyses residuals in Eigen-vector decomposition of the complete facial images [13], the second category of geometric algorithms analyses specific facial features [14]. To assess the performance of all these facial recognition algorithms, several databases have been created over the years. While some data sets with images from less than 100 subjects were designed to capture the challenges in imaging angles, facial expressions and pose, larger data sets with more than 100,000 images from over 200 subjects have been designed to address the robustness constraints of automated algorithms to variations in image qualities. In this work we analyse 3 major categories of databases, based on the number of imaged subjects and the performances of well-known methods on these databases for facial and expression recognition tasks. We observe the evolution of automated algorithms from facial recognition to expression analysis, and from recognition tasks in controlled facial images to information fusion from uncontrolled video frames.

The organization of this paper is as follows. In Section 2, the categories of facial databases are presented and their evolution is discussed. In Section 3, the facial and expression recognition methods are discussed. In Section 4 concluding remarks and discussions are presented.

## 2. FACIAL RECOGNITION DATABASES

An essential part of the constant enhancements made in the field of automated facial and expression recognition has been the collection of facial databases for benchmarking purposes. Since the 1990s there has been a drive in developing new methods for automatic face recognition as a result of the significant advances in computer and sensor technology [3-8]. Currently, there are several databases used for facial recognition which vary in size, pose, expressions, lighting conditions, occlusions and the number of imaged subjects. The earliest facial databases mostly consisted of frontal images, such as the local data set acquired from 115 subjects at Brown University used in the early works in 1987 [2]. From the year 2000 and onwards, the facial databases were seen to capture the variations in pose, lighting, imaging angles, ethnicity, gender and facial expressions [4]. Some of the most recent databases capture the variations in image sizes, compression, occlusions and are gathered from varied sources such as social media and internet [15].

Over the years, most of the well-known facial recognition algorithms have reported their performances on the databases from: AT&T Laboratories Cambridge (formerly 'The ORL Database of Faces') [16], Facial Recognition Technology (FERET) [17], Facial Database from visions Group Essex [18], Cohn Kande AU-Coded Facial Expression Database (FE) [4], NIST Mug shot Database[19], Extended Multi Modal Verification for Teleservices and Security applications (XM2VTS) Database [20], AR Face Database from Ohio [21], Yale Face Database [22], Caltech Faces [23] and Japanese Female Facial Expression (JAFFE) Database [24]. Table 1 categorizes most of the well-known facial databases into 3 categories based on the number of imaged subjects. Databases that contain images from more than 200 subjects (persons) are classified as database category 1, while the databases with images from 100-200 subjects and less than 100 subjects are classified as category 2 and category 3, respectively. The details of each database are provided for each database. For instance, the AR Face Database [21], which is very well-analysed in existing literature, belongs to database category 2 since it contains a variety of images from 126 subjects (70 men, 56 women) that represent variations in expression, illumination and occlusions. This database contains over 4000 color frontal images collected over two sessions per person on 2 separate days. The diversity of images allows for it to be used by several methodologies that focus on robust facial feature detection regardless of the extent of facial occlusions due to sunglasses and scarf. Figure 1 demonstrates the variations in facial occlusions in the sample images from the AR Face Database [21].

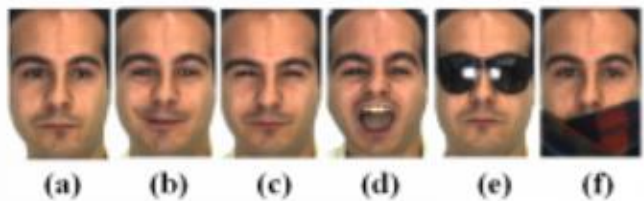

Figure 1: The six images acquired in the first session for one of the subjects in the AR face database [21] taken from the front in similar lighting conditions.

From the 3 categories of Facial Databases presented in Table 1, we observe that databases belonging to category 3 are useful for testing new methodologies on a controlled image set, while the databases from category 2 and 1 are useful for expansive parameterization of existing methods to cater to additional variations imposed by a large number of imaged subjects, imaging conditions and image formats. Also, we observe that while the early databases were focused on facial detection for subject identification, the more recent databases are geared more towards capturing the variations in imaging modalities, facial expressions, and obscurities due to makeup. Some of the latest facial databases, not shown in Table 1, are as follows:
   a. Labelled Wikipedia Faces (LWF) [25] has mined images from over 0.5 million biographic entries from the Wikipedia Living People entries and it contains 8500 faces from 1500 subjects;
   b. YouTube Faces Database (YFD) [26] contains 3425 videos of 1595 different subjects (2.15 videos per subject) with video clips ranging from 48-6070 frames. This dataset was created to provide a collection of videos and labels for subject identification from videos and benchmarking video pair-matching techniques.
   c. YouTube Makeup Dataset (YMD) [27] contains images from 151 subjects (Caucasian females) from YouTube makeup tutorials before and after subtle to heavy makeup is applied. 4 shots are taken for each subject (2 shots before and 2 shots after makeup is applied). This database has steady illumination but it demonstrates the challenges in facial recognition due to makeup alterations.
   d. Indian Movie Face Database (IMFD) [28] contains 34512 images from 100 Indian actors collected from about 100 videos and cropped to include variations in pose, expression, lighting, resolution, occlusions and makeup.

TABLE 1: CATEGORIZATION OF FACIAL DATABASES

| Category 1: Subjects above 200 | Category 2: 100-200 subjects | Category 3: Below 100 subjects |
|---|---|---|
| The BJUT-3D database [29]<br>• 2000 images,<br>• 500 distinct subjects with 4 images per subject,<br>• Variations in expressions. | Cohn- Kanade AU-Coded Facial Expression Database (FE) [4]<br>• Facial action units (AUs) and emotion-specified expressions,<br>• Version 1: 486 sequences from 97 subjects, | The Gavab DB database [30]<br>• 549 3-D images/facial surfaces,<br>• 61 subjects (45 male, 16 female),<br>• 9 images per subject,<br>• Variations in imaging angle, pose, expression. |
| | - Up to 100 university students of age range 18-30 years,<br>- Subjects were 65% female, 15% African-American, 3% Asian or Latino. | |
| The XM2VTS database [20]<br>- 295 distinct subjects,<br>- 4 recordings per subject,<br>- Speaking head shot and a Rotating head shot,<br>- High quality color images, 32 KHz 16-bit sound files, video sequences and a 3D model. | The Basel Face Model (BFM) database [31]<br>- 200 images,<br>- 100 male, 100 female subjects,<br>- Variations in pose, illumination,<br>- Generative 3D shape model covering the face surface from ear to ear and a high quality texture model. | The EURECOM Kinect Face database [32]<br>- 427 images,<br>- 52 subjects,<br>- Variations in lighting, occlusion, and 9 states of Facial Expressions. |
| The FRGC database [33]<br>- 50,000 images,<br>- 4,003 subjects,<br>- Uncontrolled and Controlled settings,<br>- Controlled Settings:<br>  - Controlled lighting conditions<br>  - 2 expressions (smiling and neutral). | The UMB database [34]<br>- 1473 total acquisitions,<br>- 143 subjects,<br>- 4 facial expressions,<br>- 3-D and 2-D images,<br>- Facial occlusions include scarves, hats, hands, eyeglasses and other. | AT&T Laboratories Cambridge (formerly 'The ORL Database of Faces) [16]<br>- 400 images,<br>- 40 subjects,<br>- Variations in lighting, occlusions, facial expression.<br>- 92x112 pixels resolution. |
| FiA "Face-in-Action" Dataset [35]<br>- 20-second videos of face data,<br>- 180 participants mimicking a passport checking scenario,<br>- Controlled, indoor environment and open, outdoor environment. | BU-3DFE (Binghamton University 3D Facial Expression) [36]<br>- 2500 images,<br>- 100 subjects,<br>- Variations in expressions,<br>- Ethnic/ racial ancestries. | Japanese Female Facial Expression (JAFFE) Database [JAFFE]<br>- 213 images,<br>- 7 facial expressions (6 basic facial expressions + 1 neutral),<br>- 10 Japanese female models. Each image rated on 6 emotion adjectives by 60 Japanese subjects. |
| ND-2006 Dataset [37]<br>- 13,450 images,<br>- 6 expressions (Neutral, Happiness, Sadness, Surprise, Disgust, and Other).<br>- 888 subjects,<br>- Upto 63 images per subject. | The CASIA-3D Face V1 database [38]<br>- 4624 images,<br>- 123 subjects,<br>- Variations in expressions, lighting and pose. | Yale Face Database [22]<br>- 165 grayscale images.<br>- 15 individuals, 11 images per subject.<br>- One image per different facial expression.<br>- Configurations: center-light, w/glasses, happy, left-light, w/no glasses, normal, right-light, sad, sleepy, surprised, and wink. |
| York 3D face database [39]<br>- Over 5000 images,<br>- 350 subjects,<br>- Controlled Setting with variations in lighting, pose, expression. | AR Face Database, Ohio State University [21]<br>- 4,000 color images,<br>- 126 subjects' faces (70 men and 56 women),<br>- Images feature frontal view faces,<br>- Variations facial expressions, illumination conditions, and occlusions (sun glasses and scarf). | MIT CBCL-Face Recognition Database [40]<br>- 10 subjects.<br>- 2 training sets:<br>  1. High resolution pictures, including frontal, half-profile and profile view.<br>  2. Synthetic images (324/subject) rendered from 3D head models of the 10 subjects.<br>- Test set consists of 200 images per subject. |

| NIST Mugshot Identification Database [19]<br>• 1573 individuals (cases) 1495 male and 78 female,<br>• Front and side (profile) views,<br>• 131 cases with two or more front views and 1418 with only one front view,<br>• 89 cases with two or more profiles and 1268 with only one profile,<br>• 27 cases with two or more fronts and one profile,<br>• 1217 cases with only one front and one profile. | The FRAV 3D database [41]<br>• 109 subjects (75 men, 34 women),<br>• 32 color captures per subject,<br>• 320 x 240 pixel resolution,<br>• 12 frontal images, 4 15°-turned images, 4 30°-turned images. 4 images with gestures, 4 images with occluded face features and 4 frontal images with a change of illumination. | The Large MPI Facial Expression database [42]<br>• 19 German subjects (10 female, 9 male),<br>• 55 Expressions,<br>• Facial expressions available in 3 repetitions, in 2 intensities and in 3 different camera angles. |
|---|---|---|
| Color FERET Database [17]<br>• 1564 sets of images,<br>• Total of 14,126 images,<br>• 1199 individuals and 365 duplicate sets of images. | Texas-3D Face database [43]<br>• 1149 images<br>• 105 subjects,<br>• Variations in expressions, gender, ethnicity | BioID Face DB, Switzerland [50]<br>• 1521 gray level images,<br>• 384x286 pixel resolution,<br>• 23 test subjects,<br>• Comparison reasons manually set eye positions. |
| CMU Multi-PIE Face Database [44]<br>• More than 750,000 images,<br>• 337 people recorded in up to four sessions,<br>• Subjects imaged under 15 view points and 19 illumination conditions,<br>• Variations in expressions,<br>• High resolution frontal images. | The Bosphorus database [45]<br>• 4666 total acquisitions,<br>• 105 subjects,<br>• 35 facial expressions from 6 basic emotions and poses,<br>• Occluded images. | Sheffield Face (Previously UMIST) database [46]<br>• 564 images,<br>• 20 subjects,<br>• Variety of poses from profile to frontal views,<br>• Variety in race, gender, appearance,<br>• 220 x 220 pixels resolution in 256 shades of grey. |
| Face Recognition Data, University of Essex, UK [18]<br>• 395 subjects (male and female),<br>• 20 images per subject,<br>• Subjects with various ethnicities,<br>• Majority of individuals age in 18-20 years,<br>• Some individuals wearing glasses and beards. | Equinox Face Data Set [47]<br>• 340 subjects, 3 different expressions,<br>• Indoor and outdoor images,<br>• For each subject images simultaneously acquired in visible, Short Wave Infra-Red (SWIR), Mid Wave Infra-Red (MWIR) and Long Wave Infra-Red (LWIR),<br>• 40 frame contiguous sequence. | Caltech Faces [23]<br>• 450 face images,<br>• 27 subjects,<br>• 896 x 592 pixels resolution,<br>• Variation in lighting, expressions, backgrounds. |
| FaceScrub [15]<br>• 107,818 face images of 530 celebrities (265 male, 265 female),<br>• 200 images per subject,<br>• Images were retrieved from the Internet (uncontrolled conditions). | SCFace- Surveillance Cameras Faces Database [48]<br>• 4160 static images (in visible and infrared spectrum),<br>• 130 subjects,<br>• Images captured in uncontrolled indoor conditions. | Indian Face Database [49]<br>• 11 images from 40 distinct subjects,<br>• Homogeneous background. Upright and frontal position images,<br>• 640x480 pixels resolution,<br>• Emotions: neutral, smile, laughter, sad/disgust. |

## 3. FACIAL AND EXPRESSION RECOGNITION METHODS

Several algorithms have been developed till date in the pursuit of improving the state-of-art in automated facial recognition. While the earlier methods focused on facial and expression analysis from images, recent methods have focused on video-based facial tracking. All the facial detection algorithms developed so far can be broadly classified into two categories. The first category of methods analyse the holistic faces and rely on residual images after Eigen-face decomposition for recognition tasks [3]. This category of methods, although computationally fast, are less adaptive to variations in pose, expression and image quality. The second category of geometric methods involve automated extraction of facial parts also known as Facial Action Units (FAUs) to compute relative distances between FAUs and their relative locations from reference points for facial and expression identification tasks [4]. This category of methods can auto-tune to capture facial expressions in motion-based images and pose variations. However, such methods require intensive training and generally have high computation time complexities [60].

Figure 2 shows the first category Eigen-face decomposition method described in [3] that estimates Eigen-vectors corresponding to a set of holistic facial images and generates a facial signature matrix that can be further modified to identify the subjects in the images in spite of occlusions, makeup and distortions [60]. Figure 3 shows the second category method of automatically extracting FAUs for expression recognition tasks. Both these methods have been demonstrated on images from the AT&T (ORL) Database [16].

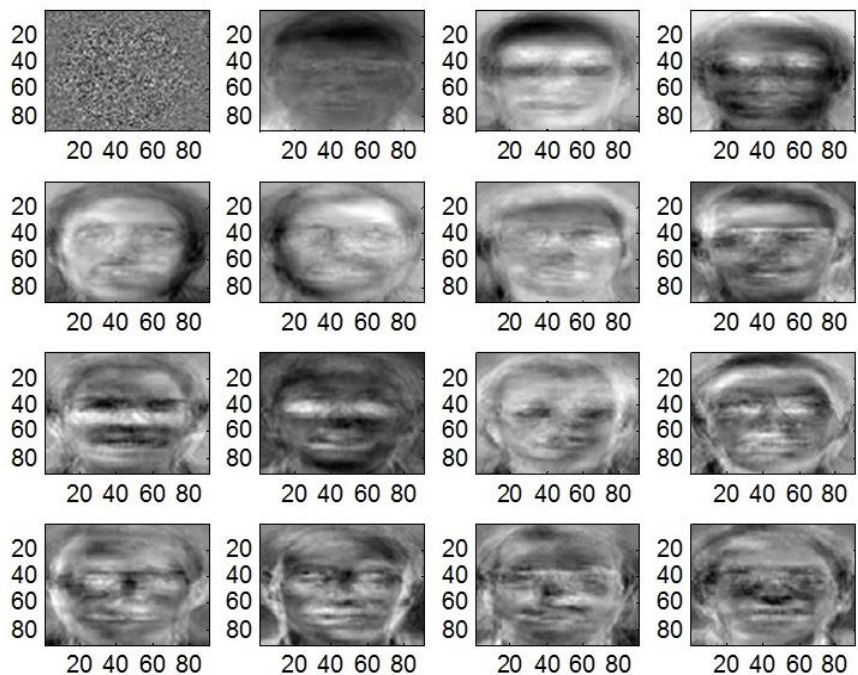

Figure 2: Example of Eigen-Face estimation using holistic facial image. The top-left image represent the averaged image or the 0th Eigen-vector. The 1st to 15th Eigen-vectors of the first image from the data base are shown thereafter.

One of the most robust algorithms for facial region detection in images is the Viola-Jones method [62] that involves Haar feature selection followed by creation of an integral image, AdaBoost classifier training and using cascaded classifiers to identify facial objects. Most of the second category of geometric methods involve the use of FAU detectors followed by classification strategies for binary or multi-class facial and expression classification tasks. Figure 4 shows an example of classification tasks involved in these FAU-based/feature-based facial recognition methods. The two classes of images have been created using images from the AT&T (ORL) database [16].

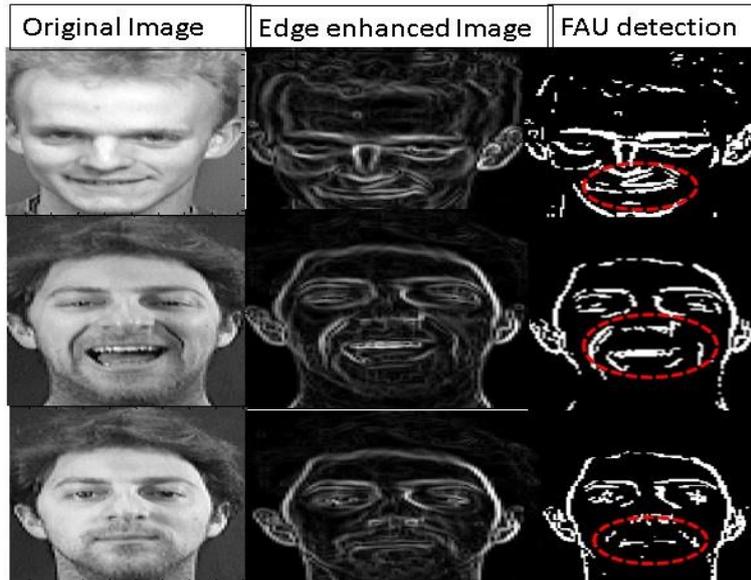

Figure 3: Example of image filtering for edge enhancement followed by automated detection of facial regions corresponding to expressions. The red circle depicts the region under analysis for "happy" expression recognition.

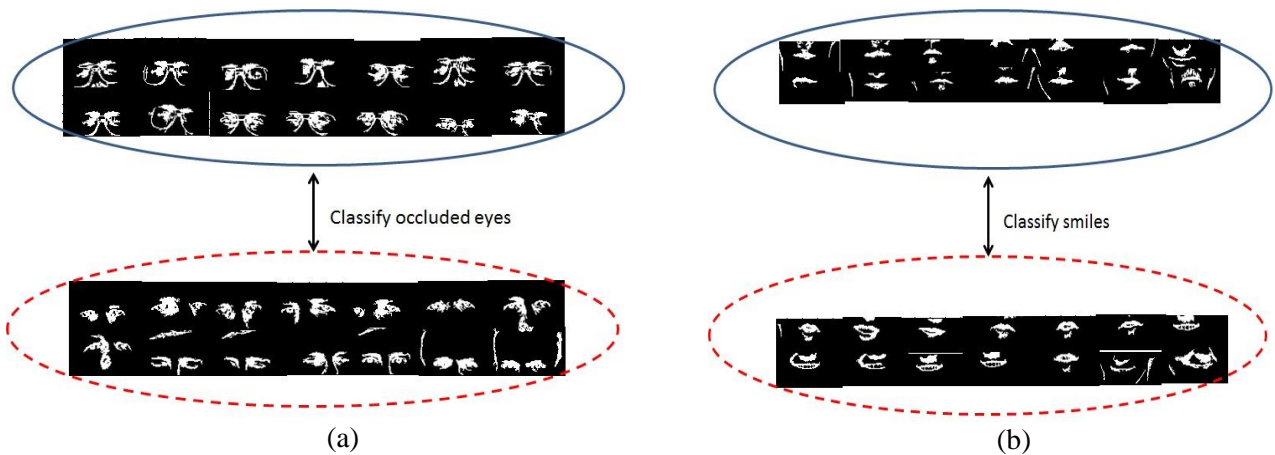

(a)                                                                 (b)

Figure 4: Example of classification tasks for facial occlusion and expression recognition. (a) Represents 2 classes of FAUs corresponding to "eye" regions in faces with glasses (occlusions) and without glasses. (b) Represents 2 classes of FAUs with and without a "smile", for "happy" emotion recognition. A typical occlusion and emotion recognition task would involve classifier training and automated separation of these two classes of facial images.

In Table 2, the well-known methods developed for facial recognition are chronologically presented. Table 2 includes the methods, features extracted for recognition tasks, the database used, the choice of classifiers and the facial recognition rates. The six images of a single session of a subject in the AR face database [21] shown in Figure 1 is used for assessing facial recognition performance of Jia et. al. [58] in Table 2. The (*) symbol indicates that the same subject's duplicate picture in the second session comprises of the training/test dataset. We observe that the early methods focused on the facial pixels as features for facial recognition followed by classification tasks. More recent methods consider splitting the images into non-overlapping regions followed by image transformation techniques for robustness to image occlusions. Classifiers such as Hidden Markov models, linear models, support vector machines and probabilistic models have been extensively used by the geometric methods [63].

## TABLE 2. CHRONOLOGICAL ORDER OF WELL-KNOWN FACIAL RECOGNITION METHODS

| Reference (Year) | Method and Features | Classification | Database | Data Composition | Performance (Recognition Rate) (%) | | | Focus |
|---|---|---|---|---|---|---|---|---|
| 1. Tan et. al. [51] 2006 | The training and test faces are disintegrated into sub-blocks followed by Self Organizing Maps (SOM) embedding using a Partial Distance (PD) metric and selecting the smallest distance face as the true identity. | Part-based nonmetric distance learning method that partitions facial images into non-overlapping blocks, computes PD and performs SOM embedding to recognize identity of a test face. | i)AT&T (ORL) [16] ii) AR Face Database [21] | **Training:** i)700 in AR dataset ii)5 images/ subject in ORL dataset **Testing:** i)1900 in AR dataset ii)5 images/subject in ORL dataset **Resized Image Resolution:** AR Face-[66x48] AT&T(ORL)-[112x92] | AR Face: 97% ORL Dataset:74.6% | | | Recognition with occlusion |
| 2. Liu et. al. [52] 2006 | Images are represented by similarity features to reference set for a relative matching strategy. | Generalization of Kernel discriminant Analysis (KDA), Linear Discriminant Analysis (LDA) | FERET database [17] | **Training:** 1002 front view face images selected from training **Testing:** FA has 1196 subjects and the FB set has 1195 subjects. | 97.41% | | | Handles nonlinear variations, especially occlusion |
| 3. Oh et. al. [53] Kim et. al. [54] (2006-2007) | Two phase method: Occlusion detection phase followed by Selective local non-negative matrix factorization phase. Each face is divided into non-overlapping patches represented by principal component analysis (PCA). The non-occluded patches are used for classification in PCA space. Occlusions are detected by combined k-nearest neighbor (kNN) and 1-nearest Neighbor (NN) threshold classifier. | Partial matching of non-occluded image parts in the PCA patch space performed by following 3 methods. a) Using projection of each image row in 2-D PCS subspace. b) Partitioning each face to 6 non-overlapping blocks (3 on left, 3 on right) and projecting them on PCA sub-space. c) Image transformation followed by projection of each row in PCA sub-space. | AR-face Database (76 men, 59 women) [21] | **Training:** i) 35 sunglass images. ii)35 scarf images from 20 men and 15 women. **Testing:** i)100 sunglass images ii)100 scarf images. | types of occlusions | scarf | sun-glasses | Facial recognition with occlusions |
| | | | | | Method a) | 99% | 98% | |
| | | | | | Method b) | 98% | 96% | |
| | | | | | Method c) | 98% | 98% | |
| 4. Zhang et. al. [55] 2007 | Each face is represented by local Gabor binary patterns (LGBP) converted into a feature vector. Using Kullback-Leibler Divergence (KLD) distance between LGBP features, the probability of occlusions is estimated. Occlusion probability is the weight for feature matching. | Thresholded occlusion probability computed using the KLD distance metric is used for classification. | AR face database [21] | **Training:** 50, [80×88] resized by position of eyes randomly chosen subjects with neutral expression. **Testing:** Synthetic occlusions added to neutral, smiling, screaming and angry faces. | Imaged Session | Sunglasses | Scarf | Recognition with occlusion |
| | | | | | 1 | 84% | 100% | |
| | | | | | 2 | 80% | 96% | |
| 5. Lin et. al. [56] 2008 | Posterior union model decision-based neural network (PUDBNN). Posterior union model (PUM) ignores severely mismatched features and focusses on matched local features. Partial distortions are simulated by adding sunglasses, beards (male) and scarf (female). | Local Features comprise of 3 level db4 wavelet transform. Each face represented by 25 coefficients as local features. Neural Networks classifier performs facial recognition decision. | XM2VTS [20] and AT&T (ORL) Databases [16]. | **XM2VTS:** 100 subjects selected randomly with 4 images per subject. **Training:** 3 images per subject. **Testing:** 1 image per subject. **AT&T:** 40 subjects with 10 images per subject. **Training:** 5 images per subject. **Testing:** 5 images per subject. | Database | Average Recognition Rate | | Recognition with occlusion and unknown partial distortions. |
| | | | | | ORL | 83.5% | | |
| | | | | | XM2VTS | 82.4% | | |
| 6. Guo, et al. [57] Jia et. al. [58] (2001-2009) | Partial Support Vector Machines (PSVM) criterion is introduced to work with missing feature components. Features comprise of facial pixel values. The goal is to minimize probability of overlap between most probable values of samples in any class. | PSVM classification enables training and testing for facial recognition on occluded and non-occluded faces. The occlusions are artificially added to training images by overlaying [sxs] pixels in random locations, where, s=0,3,6,9,12. | AR face database [21] and FRGC version 2 dataset [33] | **AR Database:** 1200 images. Images are cropped and resized to [29×21] pixels **FRGC version 2 dataset:** 800 images. Images are cropped and resized to [30×26] pixels. Cropping and resizing operation are performed to align faces with respect to location of eyes, nose and mouth. | Images in Training sets (refer to Figure 1) | Images in Testing sets (refer to Figure 1) | Result | Recognition with random occlusions. |
| | | | | | [a, e, f] | [b, c, d] | 88.9% | |
| | | | | | [a*, e*, f*] | [b*, c*, d*] | 90.8% | |
| | | | | | [a, b, c, e, f] | [d] | 88.2% | |
| | | | | | [a, b, c, e, f] | [d*] | 58.8% | |
| | | | | | [a, b, c, e, f, a*,b*,c*c, e*, f*] | [d, d*] | 83.5% | |
| 7. Lin et. al. [59] 2009 | A similarity-based metric is introduced to probability metric in posterior union model (PUM) for reliable recognition of local images to improve mismatch robustness for facial recognition. Only one or few training images needed. Each face is partitioned into 16 non-overlapping local images, 5 scales x4 | Gaussian Mixture Model (MM) classifiers are trained on a large feature vector from few images. Each image has 11,520 feature vector. Testing images are clean and corrupted images with partial distortions by adding 4 types of occlusions: sunglasses, | XM2VTS [26] and AR face database [21] | **XM2VTS:** 100 subjects chosen randomly 4 images/ subjects of which one or two for training and remaining for testing with corrupted occlusions. **AR Face Database:** 4 images /subject of which one or two for testing and remaining for testing with corrupted occlusions. Each face resized to [96x96] | Database | 1 Training Image (TI) | 2 TI | 4 TI | Facial recognition with partial simulated occlusions. |
| | | | | | AR | 79.5% | NA | 91.5% | |
| | | | | | XM2VTS | 88.3% | 96.8% | NA | |

| | orientations of Gabor filter to each image and down-sampling coefficients by 4. | beard/scarf, sunglasses and beard/scarf, hands. | | pixels. | | | | |
|---|---|---|---|---|---|---|---|---|
| 8. Wright et. al. [60] 2009 | Sparse representation framework based on $l_1$-minimization is shown to be more useful for facial classification than feature selection. Down sampled images, random projections, Eigen faces and Laplacian faces are equally important features as long as the dimension of the feature set surpasses a threshold. | Sparse Representation based Classification (SRC) minimizes $l_1$ norm by primal-dual algorithm for linear programming. Partial and full face features are computed to ensure number of training samples >=1,207. | AR Face Database [21] and Extended Yale B database [22] | **AR Face Database:** 50 male, 50 female subjects. 14 images per subject with only illumination change and expressions were selected. **Training:** 7 images from Session 1 per subject **Testing:** 7 images from Session 2 per subject. **Extended Yale B Database:** 38 subjects. Controlled lighting. **Training:** 32 images per subject randomly selected. **Testing:** Remaining 32 images per subject. | Occlusion region | Recognition Rate | Feature extraction from facial images and robustness to occlusions. |
| | | | | | Nose | 87.3% | |
| | | | | | Right Eye | 93.7% | |
| | | | | | Mouth and Chin | 98.3% | |
| | | | | | Percentage Occlusion 50% | 100% | |
| | | | | | Sunglasses | 87-97.5% | |
| | | | | | Scarves | 59.3-93.5% | |
| 9. RoyChowdhury et. al. [61] (2015) | Bilinear Convolutional Neural Networks (B-CNN) is applied for facial recognition in large public data sets with pose variability. 4 sets of features are evaluated: traditional features (eyes, nose, mouth, and eyebrows), features correlated to accessories, features correlated with hair, and features correlated with background. | B-CNN model is a directed acyclic graph (DAG) with backpropagation step to learn inter neuron weights and biases. Image labels ae used or training and testing. | FaceScrub Database [15] | **Training:** 203 images/subject (513 subjects)+external data **Validation:** One third of images in each class. | Method | Recognition Rate | Facial recognition from a large public domain data set to identify features that correlate with background, accessories and facial occlusions. |
| | | | | | No fine tuning | 31.2% | |
| | | | | | After fine tuning | 52.5-58.8% | |

In Table 3, the chronological development in facial expression recognition methods is shown. Here, we observe that the early methods involved extraction of facial action units (AUs) from images for automated expression recognition algorithms for expressions such as anger, disgust, sadness, happy, surprise. Recent methods focus on expression analysis rom video recordings and variations in training and test data sets.

Some methods that perform automated facial and expression recognition on the recent data bases mentioned in Section 2 are as follows:
   a. Dago-Cass et. al. [73] (2011): Gender classification is performed using appearance based, feature based (Gabor coefficients and local binary patterns) descriptors with linear SVM and discriminant analysis. 60-94% recognition rates are achieved on the LFW database [25].
   b. Wolf et. al. [26] (2011): Facial recognition in unconstrained video with matched background similarity is performed. The SVM classifier is used on a data set comprising of 1 video for 591 subjects, 2 videos for 471 subjects, 3 videos for 307 subjects, 4 videos for 167 subjects, 5 videos for 51 subjects and 6 videos for 8 subjects. 50-77% recognition rates are achieved on the YF database [26].
   c. Chen et. al. [27] (2013): Shape, texture, color of images with and without makeup are used for facial recognition tasks (1484 features). Geometric features corresponding to certain regions of the face (such as eyes, mouth) are extracted followed by classification by SVM and AdaBoost. Recognition rates for facial AUs and full face lie in the range 58-91% on the YM database [27].
   d. Beham et. al. [74] (2014): Dictionary-based approach to extract features and perform k-means clustering with sparse representation. The images from the IMFD [28], CMU Pie [44] and Extended Yale B [22] datasets have variations in illumination, expression and controlled/uncontrolled settings. The expressions are captured by rotation invariant local binary patterns and histogram of gradients.
   Sarode et. al. [75] (2014): Automated facial recognition from video frames with variations in pose and appearance is performed. Modified Genetic Algorithm (GA) based transfer vectors are used for generating facial features from different poses.

Classification is performed by k-nearest neighbor (k-NN) and discriminant analysis on the FERET database [17] and an unconstrained database created similar to IMFD [28]. Recognition rate ranges from 12-91.81% for FERET Database [17] and 6.55-25.32% for the unconstrained database.

Kumar et. al. [76] (2014): Sparse framework with $l_1$-minimization is used for facial recognition in the IMFD [28] database for robustness to age, illumination, pose, expression, lighting and storage limitations in images extracted from videos. Two kinds of features are extracted: Scale invariant feature transform (SIFT) and local binary patterns. The features are reduced by performing principal component analysis (PCA) followed by supervised classification using k-NN and sparse representation classifier. For images in IMFD [28] (with at least 200 images per subject) from each of the 100 subjects, 100 images are used for training and the rest are used for testing. Recognition rates in the range 55-69% are achieved.

## 4. CONCLUSIONS AND DISCUSSION

Over the past few decades facial and expression recognition topics have been significantly analyzed and there have been significant changes in the innovation trends. While early methods focused on semi-automated facial recognition, the later methods shifted focus to develop fully automated facial recognition methods that are robust to pose, illumination, imaging/lighting limitations, occlusions and expressions. To facilitate comparative assessment between methods, several public data bases evolved to capture the limitations of automated facial recognition. This facilitated significant analysis of robust methods that extracted holistic facial features and geometric action units (AUs) from faces for facial recognition in the event of occlusions. Thereafter, the focus of methods shifted to automated facial expression recognition, where expressions could vary as neutral, happy, sad, surprise, anger and disgust. This caused a shift in the trend of the public databases that began to be focused on subject level expression detection.

Automated age, ethnicity and gender detection methods and databases were also developed, but they were not as significantly analysed as the automated expression recognition problem. The most recent trends have further moved towards automated facial and expression recognition from images that vary over time (video recordings) and images that vary over space (collected from the internet). This caused a shift of focus to methods that are independent of image storage and compression limitations. Some of the well-known recent databases collect images from videos in the internet and they represent a wide variation in image storage and quality [15] [25-28]. This motivates further research into scalable cloud-based methods that can extract features from large databases and correlate them with facial recognition tasks. Thus, future trends may involve automated robust facial/expression recognition in video streams that vary over space (e.g. auto-tagging of subjects as they age from video recordings gathered over the internet).

In this work we categorize the well-known facial and expression recognition databases based on the number of subjects imaged. While the databased with images from lesser number of subjects can be useful for applications involving facial and expression recognition in limited/controlled settings, the databases with large number of subjects can be useful for background equalization, and recognition tasks on images with uncontrolled imaging parameters. Further, the analysis of facial recognition and facial expression recognition methods shows that facial expression and occlusions pose a bigger challenge to robust automated facial and expression recognition methods than gender, ethnicity and age of subjects.

TABLE 3: CHRONOLOGICAL ORDER OF FACIAL EXPRESSION RECOGNITION METHODS

| Reference (Year) | Method | Features and Classification | Database | Data Composition | Performance (Recognition Rate) | Properties |
|---|---|---|---|---|---|---|
| 1. Tian et. al. [4] (2001) | Automatic Face Analysis (AFA) system based on permanent facial features (brows, eyes, mouth) and transient facial features (facial furrow depths) in a frontal-view face image sequences. Recognition of 16-30 facial action units (AUs). | 2 artificial neural networks (ANN) for: i) upper face recognition ii) lower face recognition. **Permanent Features:** Optical Flow, Gabor Wavelets and Multi-State Models. **Transient Features**: Canny Edge Detection | Cohn-Kanade [4] and Ekman-Hager [5] | Upper Face: 50 sample sequences from 14 subjects performing 7 AUs. Lower Face: 63 sample sequences from 32 subjects performing 11 AUs. | Average Recognition rate upper face 96.4% AUs (95.4% excluding neutral expression) and Lower face 96.7% AUs (95.6% excluding neutral expression) Computation time: Less than 1 second per frame pair | Real-time system recognizes posed expressions of happiness, surprise, anger, fear. Motion is invariant to scaling. Uses facial feature tracker to reduce processing time. |
| 2. Tan et. al. [64] 2005 | Self-organizing map (SOM) is used to learn the subspace occupied by each facial image. Next, soft k-nearest neighbor (kNN) classifier is used to recognize facial identity. | Image divided into non-overlapping blocks of equal size. Local face features extracted from the image sub-blocks, Eigen-faces and Gabor filtered sub-images. The k-NN classifier outputs a confidence vector with highest degree of support for the most probable facial feature vector | AR database [21] and FERET database [17] | **AR Database:** 100 subjects (50 male, 50 female). **Training:** Neutral expression from 100 subjects **Testing:** Smile, anger, scream expression from 100 subjects. **FERET Database**: 1196 subjects (1 image/subject) Probed training and testing | Facial Recognition: 64-100% With variations in expression, and occlusions. | Automated facial recognition with partial occlusions and varying expression |
| 3. Faltemier et. al. [37] 2007 | Multi-Instance Enrollment Representation for 3-D Face Recognition. Using 3-D shape of frontal view face, up to 28 regions are detected and aligned using iterative closest point registration (ICP) algorithm. | Accuracy of nose-tip image determines expression recognition rate. ICP iterates till root mean square (RMS) scores<0.0001 or 100 iterations. For good initial alignment iterations needed=30. | ND-2006 data corpus (3D) [37] | 888 subjects with up to 63 images/subject. 13,450 images containing 6 different types of expressions {Neutral: 9899 images, Happiness: 2230 images, Sadness: 180 images, Surprise: 389 images, Disgust: 205 images, Other: 557 images}. | Expressions / Recognition Rate<br>Neutral+Happy+Sad+Surprise: 98%<br>Neutral: 82.8%<br>Happy: 85.1%<br>Sad: 91%<br>Surprise: 89.8% | Expression Recognition. Heterogeneous expression sampling improves recognition rate. |
| 4. Gundimada et. al. [65] 2009 | A modular Kernel Eigen Spaces-based approach using feature maps extracted from visual and thermal images. After each training image is modularized, a kernel matrix is estimated for each vectorized module followed by KPCA for module weight determination. | Phase congruency features are extracted for each test face followed by minimum-distance metric based classification. Gaussian Radial Basis Function kernel and polynomial kernel are used. | AR face database [21] and Equinox database [47] | **AR data base:** 40 subjects randomly chosen. 13 images/subject. **Training:** 3 images /person, neutral expression. **Testing:** 10 images/ individual Equinox Face Database: Longwave infrared and visual spectrum face for 34 subjects. 15 images/subject. | Recognition type / Recognition Rate<br>Occlusion: 94.5%<br>Facial recognition: 83%<br>Decision Fusion: 87.4% | Robust to illumination variations, partial occlusions, expression variations, and variations due to temperature changes that affect the visual and thermal face recognition techniques |
| 5. Xue et. al. [66] 2014 | Face images are transformed into color spaces by concatenating their component vectors. Facial expression recognition is achieved by utilizing Fisher Linear Discriminant (FLD). Uncorrelated color space (UCS), discriminant color space (DCS) are derived for each face. | FLD is used to extract expression features followed by classification using nearest neighbor (NN) classifier. Subject-independent (**IND**) and subject-dependent (**DEP**) experiments are conducted. | Oulu-CASIA NIR&VIS facial expression database [38] and Curtin Faces database [67] | **Oulu-CASIA NIR&VIS :** 80 subjects (73.8% male). 6 expressions. Images are frame sequences from videos. The first 9 images of each sequence are ignored. The selected 6,059 images are aligned by position of eyes and mouth. **IND:** Training on first 40 subjects testing on remaining 40 subjects. **DEP:** Training on images from first half of video sequences and testing on the remaining half video frames. **Curtin Faces Database:** 52 subjects, 5000 images. 6 images/subject. **IND:** Training on 25 subjects, testing on 27 subjects. **DEP:** Training on | The average recognition rates on Oulu-CASIA database (%)<br><br>| Color space | Gray | RGB | DCS | UCS |<br>|---|---|---|---|---|<br>| IND | 49.5 | 49.9 | 48.6 | 53.0 |<br>| DEP | 91.3 | 91.4 | 91.7 | 92.5 |<br><br>The average recognition rates on Curtin Faces Database<br><br>| | | | | |<br>|---|---|---|---|---|<br>| IND | 42.8 | 45 | 42.7 | 47.1 |<br>| DEP | 45 | 49.6 | 49.7 | 53.7 | | Learning optimal color space for facial expression recognition |

| | | | | first 3 expressions per subject, for 52 subjects and testing on remaining 3 images/subject. | | | | |
|---|---|---|---|---|---|---|---|---|
| 6. Saeed et. al. [68] 2014 | Independence from prior neutral expression in training phase for classification of anger, disgust, fear, happiness, sadness and surprise. 8 facial points detected in a bounding rectangle around the face. Features represent location and shape of eyes, eyebrows and mouth. Point distribution model (PDM) to avoid shape distortions. | Geometrical features corresponding to 6 distance metrics from 8 facial points are extracted. Distances are normalized with ace width. K-Nearest-Neighbor (KNN) & Several one-versus-all (SVM) classifier used for expression classification. | Cohn Kanade datase(CK+) [8] and Binghamton University 3D dynamic Facial Expression Database (BU-4DFE) [69] | **CK+ :** 123 subjects, 593 image sequences. Each sequence starts with neutral and ends with peak expression. 327 of the sequences were labeled for the human facial expressions: 45 for anger (An); 18 for contempt (Co); 59 for disgust (Di); 25 for fear (Fe); 69 for happiness (Ha); 28 for sadness (Sa); 83 for surprise (Su) Leave-One-Out Cross Validation. **BU-4DFE:** 6 expressions and neutral expression. | Expression<br><br>An<br>Di<br>Fe<br>Ha<br>Sa<br>Su | Recognition Rate (Best) (%)<br>CK+<br>80<br>83<br>72<br>100<br>64.2<br>98.7 | BU-4DFE<br>62.7<br>59.8<br>55.9<br>88.4<br>53.4<br>93.7 | Geometrical feature based expression recognition that is robust to training on neutral expression images. |
| 7. Valstar et. al. [70] 2015 | Estimation of facial action unit (AU) intensity and occurrence with respect to manual annotations in video frames. Three tasks: detection of AU frequency, estimation of AU intensity and fully automatic AU intensity estimation. Features include 2-layer appearance features (Local Binary Gabor Patterns) and geometric features based on 49 tracked landmarks. Total 316 geometric features per video frame. | Comparative assessment of classifiers: Artificial neural networks (ANN), boosting, SVM. | BP4D Spontaneous database [71] and SEMAINE database [72] | **BP4D Database**: video recordings of young adults responding to emotion inducing tasks. 8 sessions/subject. **Training**: 41 subjects (56.1% female, 49.1% white, age 18-29). 68 sessions (75,586 images). **Testing:** 20 subjects. Includes physiological data and thermal images. 71,261 images in development and 75,726 in testing (222,573 in total). **SEMAINE:** User interaction (emotion) recordings. 49.979 frames/sec. 24 recordings. **Training:** 16 sessions. 48,000 images Development: 15 sessions, 45,000 images **Testing:** 12 sessions, 37,695 images. | Test Data<br><br>Development partition AUs in BP4D<br><br>Development partition AUs in SEMAINE<br><br>Testing Partition AUs in BP4D<br><br>Testing Partition AUs in SEMAINE | Recognition Accuracy (%)<br><br>56.9-84.4 (Geometric) 46.8-81 (Appearance)<br><br>59.1-93.2 (Geometric) 32.7-97.5 (Appearance)<br><br>54.1-72.3 (Geometric) 391.-77.4 (Appearance)<br><br>68-83.2 (Geometric) 35.7-93.8 (Appearance) | | Face Action Unit recognition in video recordings. Up to 8 emotions detected. |

Future efforts in the field of facial and expression recognition may involve identification of expression-based biometrics that can be useful for automated security, surveillance, and identification tracking tasks [77]. Robust automated facial/expression recognition can be used as personal identification systems at grocery stores, travel documentation, banking documentation, examinations and security, and also criminal tracking. Methods that can extract facial informatics from large internet-based data sets can be useful for "Quantitative face" technologies, where every aspect of the face and expression can be mathematically quantified for extremely high resolution information extraction, tracking and monitoring using facial expressions alone.

**Authors**

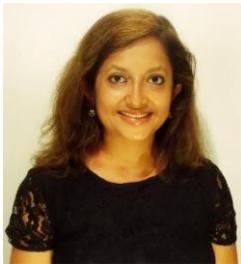

[1]**Sohini Roychowdhury** received her Ph.D. in Electrical and Computer Engineering from University of Minnesota in August 2014 and M.S. from Kansas State University in May 2010. She is currently an assistant professor in the Department of Electrical Engineering at University of Washington, Bothell. Her research interests include image processing, signal processing, pattern recognition, machine learning, artificial intelligence, low power system design and cloud computing. She is the recipient of two best paper awards, one best poster award and one best paper finalist at the Institute of Engineering and Medicine Conference (2013), IEEE Asilomar Signals, Systems and Computers Conference (2012), IEEE Student Paper Contest Alborg University (2007) and Osmosis Student Paper Contest (2006), respectively. Her online screening system for images with Diabetic Retinopathy (DReAM: Diabetic Retinopathy Analysis Using Machine learning) was featured as the Cover article of the IEEE Journal on Biomedical and Health Informatics in September 2014. She is also the winner of the Graduate School Fellowship for the year 2010 and numerous travel grants, at the University of Minnesota.

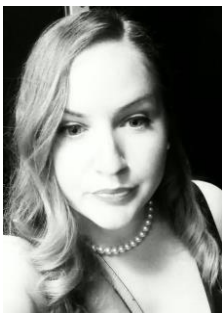

[2]**Michelle L. Emmons** Michelle Emmons received her Bachelor of Science in Electrical Engineering from the University of Washington after attending the Bothell campus in 2015. She served as an avionics technician in the United States Navy. Her military background motivated her interest in Facial and Expression Recognition as a topic of undergraduate research project.

.